# The application of GPT-4 in grading design university students' assignment and providing feedback: An exploratory study

Qian Huang, Lee Kuan Yew Centre for Innovative Cities, Singapore University of Technology and Design, Singapore. Email: qian_huang@sutd.edu.sg

Thijs Willems, Lee Kuan Yew Centre for Innovative Cities, Singapore University of Technology and Design, Singapore, Email: thijs_willems@sutd.edu.sg

Poon King Wang, Lee Kuan Yew Centre for Innovative Cities, Singapore University of Technology and Design, Email: poonkingwang@sutd.edu.sg

**Abstract**

This study aims to investigate whether GPT-4 can effectively grade assignments for design university students and provide useful feedback. In design education, assignments do not have a single correct answer and often involve solving an open-ended design problem. This subjective nature of design projects often leads to grading problems, as grades can vary between different raters, for instance instructor from engineering background or architecture background. This study employs an iterative research approach in developing a Custom GPT with the aim of achieving more reliable results and testing whether it can provide design students with constructive feedback. The findings include: First, through several rounds of iterations the *inter*-reliability between GPT and human raters reached a level that is generally accepted by educators. This indicates that by providing accurate prompts to GPT, and continuously iterating to build a Custom GPT, it can be used to effectively grade students' design assignments, serving as a reliable complement to human raters. Second, the *intra*-reliability of GPT's scoring at different times is between 0.65 and 0.78. This indicates that, with adequate instructions, a Custom GPT gives consistent results which is a precondition for grading students. As consistency and comparability are the two main rules to ensure the reliability of educational assessment, this study has looked at whether a Custom GPT can be developed that adheres to these two rules. We finish the paper by testing whether Custom GPT can provide students with useful feedback and reflecting on how educators can develop and iterate a Custom GPT to serve as a complementary rater.

## 1.Introduction

Since its inception, ChatGPT has revolutionized various industries, including education (Montenegro-Rueda et al., 2023). The potential for ChatGPT to transform educational settings has sparked significant debate and discussion (Arif, Munaf & Ul-Haque, 2023; Hwang & Chen, 2023). A growing number of educators recognizes the necessity of embracing technology to enhance learning outcomes (Sim, 2023). Universities, such as Arizona State University, have integrated ChatGPT into their teaching and research frameworks, demonstrating its practical applications[1]. Moreover, organizations and educational institutions have begun establishing guidelines for the ethical and effective use of generative AI in educational contexts (Halaweh, 2023). The advent of GPT-4 has marked a significant leap forward in AI accuracy and utility in academic contexts. For instance, GPT-4 improved translation accuracy significantly (Chen, 2023); GPT-4 demonstrated superior performance on medical competency examinations (Nori et al., 2023); and Custom GPT models, for instance, "Scholar GPT[2]" has been developed to enhance research productivity and tailor solutions to specific professional needs.

Despite these improvements, GPT-4 struggles with reasoning tasks and is often unable to consistently solve complex problems (Arkoudas, 2023). GPT-4, while improved from earlier versions, is still prone to hallucination and fabrication in responses (Currie, 2023). Concerns also persist regarding the potential erosion of students' critical thinking and problem-solving skills due to over-reliance on AI (Luckin, 2017). Ethical considerations, including data privacy and inherent biases in AI algorithms, remain significant (Binns & Veale, 2023). AI systems, including ChatGPT, can inadvertently perpetuate biases present in their training data, leading to unfair outcomes for certain student groups (Weidinger et al., 2022). Therefore, continuous efforts to audit and refine AI models are essential to ensure fairness and inclusivity (Halaweh, 2023).

In light of the rapid advancements brought by GPT-4 and the development of custom GPT models, the education sector stands at a critical juncture. While these technologies offer impressive enhancements to research and pedagogy, their potential downsides cannot be overlooked. This duality presents a complex challenge for educators and policymakers: how to harness the benefits of AI tools like ChatGPT while mitigating their risks. This is not only about leveraging technology for academic excellence but also about ensuring that it serves as a tool for equitable and ethical education (Luckin, Holmes, & Griffiths, 2022). Moving forward, it becomes crucial to develop robust mechanisms for monitoring AI's impact on learning environments and to ensure that educators are well-equipped to utilize these technologies in ways that truly enhance student learning and critical thinking skills.

While the emergence of many studies discussing the impact of Gen-AI on education is important and crucial in assuring that its developments vis-a-vis learning and teaching are monitored (Hwang & Chen, 2023; Kadaruddin, 2023), we lack a robust understanding of Gen-AI in practice and how it either benefits or hampers educators and/or students. To this end, this study sets out to empirically study the extent to which a Custom GPT-4 model can assist teachers in one of their core competencies: the grading of students' assignments. Specifically, we look at how a Custom GPT can or cannot be of help in the grading of subjective and open-ended assignments, in our case design projects. While the grading of yes/no or multiple-choice questions is arguably more straightforward, assessing subjective assignments can even amongst human raters be a controversial topic with different raters giving different grades.

---

[1] https://ai.asu.edu/openAI
[2] https://chatgpt.com/g/g-kZ0eYXUe-scholar-gpt

This research addresses the critical gap by developing and empirically testing custom GPT-4 models to evaluate their effectiveness as grading assistants in a design course at a Singaporean design university. Through multiple iterations, we aim to assess and enhance the accuracy and reliability of GPT in grading tasks. This study not only seeks to validate the use of GPT for educational assessments but also provides actionable guidelines to educators on leveraging this technology effectively.  The goal of this exercise is to determine the conditions and GTP instructions under which a custom GPT can act as a reliable grader that human raters can use to complement their work. We deliberately use the word complement here, because even though we found that custom GPT-models can provide reliable and consistent assessments we think that the domain knowledge of educators remains crucial. However, assessments done by a custom GPT can serve as a useful benchmark or act as an additional evaluator if different human raters are in disagreement.

The two research questions of this study are: 1) Can GPT accurately grade students' assignments in a design course, and 2) under which conditions can a Custom GPT act as a reliable grader to serve as a complementary rater? By answering these questions, this research aims to verify through empirical studies, involving multiple iterations, whether GPT can be an accurate and reliable grading assistant. Additionally, it seeks to provide guiding principles for educators on how to achieve this reliability.

## 2.Literature

*2.1 ChatGPT in education*

ChatGPT, developed by OpenAI, is a generative pre-trained transformer that has seen widespread adoption in various sectors, including education. Its capabilities extend from providing personalized learning experiences to facilitating administrative tasks, thereby reshaping traditional educational landscapes (Halaweh, 2023; Poon, Willems, & Huang, 2024, Strzelecki, 2023; Zhu & Li, 2023).

Educators can leverage ChatGPT to assist in creating and delivering content. The AI can generate lecture materials, interactive discussions, and tailor assessments to student needs (Chung & Park, 2023; Kohnke & Zou, 2023). For instance, AI-driven tools like ChatGPT can revolutionize traditional assessment methods. They support innovative testing approaches, such as adaptive testing, which aligns the difficulty of test items with the student's ability level, potentially leading to more accurate measurements of student knowledge and skills (Sok & Heng, 2024).

These transformative capabilities in the application of GPT in education can be categorized in several key areas.  *Academic writing*: Studies have explored the potential of Chat GPT as a tool for academic editing, highlighting its abilities in error detection, writing improvement, and content generation (Tao & Zhang, 2023; Zhu & Wang, 2023).  *Personalized feedback*: Studies have explored using GPT-3.5 to generate personalized feedback for programming assignments (Wang & Yang, 2023; Wilson & Evans, 2022).  Preparing teaching material: It can be employed to design teaching contents, materials, quizzes (Anderson & Morgan, 2023). Analyzing qualitative data: Some studies have explored how ChatGPT can be used to enhance research capabilities in several fields by helping with data analysis (Wang & Liu, 2023).

Building on the versatile applications of ChatGPT in education, it becomes evident that this technology not only supports traditional educational functions but also introduces innovative evaluation mechanisms that enhance reliability and precision in assessments. The transition from AI-assisted content creation and personalized learning tools to sophisticated grading systems exemplifies how AI can bridge the gap between educational content delivery and assessment accuracy. This shift underscores the importance of integrating reliable AI tools, like ChatGPT, to maintain the integrity and fairness of educational evaluations. By leveraging AI for both teaching and assessment, educators can ensure a more cohesive and supportive learning environment that aligns teaching methods with evaluation practices.

*2.2 Reliability of Educational Assessment*

Ensuring the reliability of educational assessments is critical for valid and equitable evaluation of student performance. This is even more so the case when evaluating the accuracy of a Custom GPT rater. To evaluate the above, we can draw on two key indicators that were initially developed in the field of psychometrics and have since then been applied in fields such as the social sciences, mathematics, education, etc.: inter-rater reliability, and intra-rater reliability. The latter concerns the consistency of a single rater over time, whereas the former concerns the consistency among multiple raters. Despite the theoretical discussions on GPT's potential as a grading assistant, empirical studies remain scarce. It is imperative to bridge this gap by translating theoretical insights into practical applications and providing clear guidelines for educators.

<u>Intra-Rater Reliability</u>

Intra-rater reliability refers to the degree to which the same examiner gives consistent ratings of the same subject on different occasions. High intra-rater reliability is essential as it indicates that assessments are not influenced by external factors such as mood, environment, or temporal context, which could potentially bias the rater's judgment. For instance, a study by Ling, Mollaun, and Xi (2014) found that fatigue could negatively affect a human rater's scoring quality on constructed responses.

Intra-rater reliability is commonly measured using the intraclass correlation coefficient (ICC). An ICC above 0.70 is generally considered good, indicating strong consistency, while values between 0.60 and 0.74 are considered acceptable for normal assessments. For high-stakes assessments, an ICC or kappa value above 0.85 is recommended to ensure maximum reliability (Saxton et al., 2012).

Intraclass Correlation Coefficient (ICC) is a commonly used statistic for evaluating consistency or agreement among raters, which is, inter-reliability of multiple raters. It measures the reliability of ratings by comparing the variability of different ratings of the same subject to the total variation across all ratings and all subjects. The two-way ANOVA model is often used to estimate the ICC by treating both subjects and raters as random effects, which helps in understanding the variability attributed to different sources Cicchetti, 1994; Weir, 2005).

Inter-Rater Reliability

Inter-rater reliability is typically measured using intraclass correlation coefficient (ICC). For normal assessments, an ICC value of 0.70 to 0.80 is considered satisfactory (Cicchetti,1994). Inter-rater reliability is crucial for ensuring consistency among multiple graders. A study showed that trained raters achieved acceptable levels of inter-rater reliability (α ≥ 0.70) across all rubric categories, highlighting the importance of comprehensive rater training and clear assessment criteria (Saxton et al., 2012).

Further, Şahan and Razı (2020) explored how raters with different levels of experience approached the assessment of essays. They found that raters prioritized different aspects of essays based on their quality and that rater behaviors evolved with practice, indicating that training impacts raters differently depending on their experience. While Weigle (2002) emphasized that detailed rubrics and ongoing rater training are essential for maintaining high inter-rater reliability in language assessments.

### 3.Research methods

The background of this research takes place in a design thinking course for first-year students at a university in Singapore. The researchers observed in this course that each class had one engineering instructor and one architecture instructor. Student feedback often indicated that the advice given by the two instructors was different, and the students did not know how to balance the feedback from both teachers. Therefore, the researchers attempted to use ChatGPT-4 to grade the design assignments of 10 students. Each student submitted 2 poster designs, making a total of 20 screenshots inputted into ChatGPT-4.

In terms of the testing the reliability of ChatGPT grading, this study employed a Design-Based Research (DBR) approach, which is commonly used in educational experiments and reforms (Campanella & Penuel, 2021). Through multiple iterations and prototypes, the researchers aimed to find better ways to address the research questions. This iterative process is fundamental in DBR, allowing for continuous refinement and adaptation of strategies based on the results obtained from each phase of testing and prototypes. Figure 1 shows the process of DBR in this study.

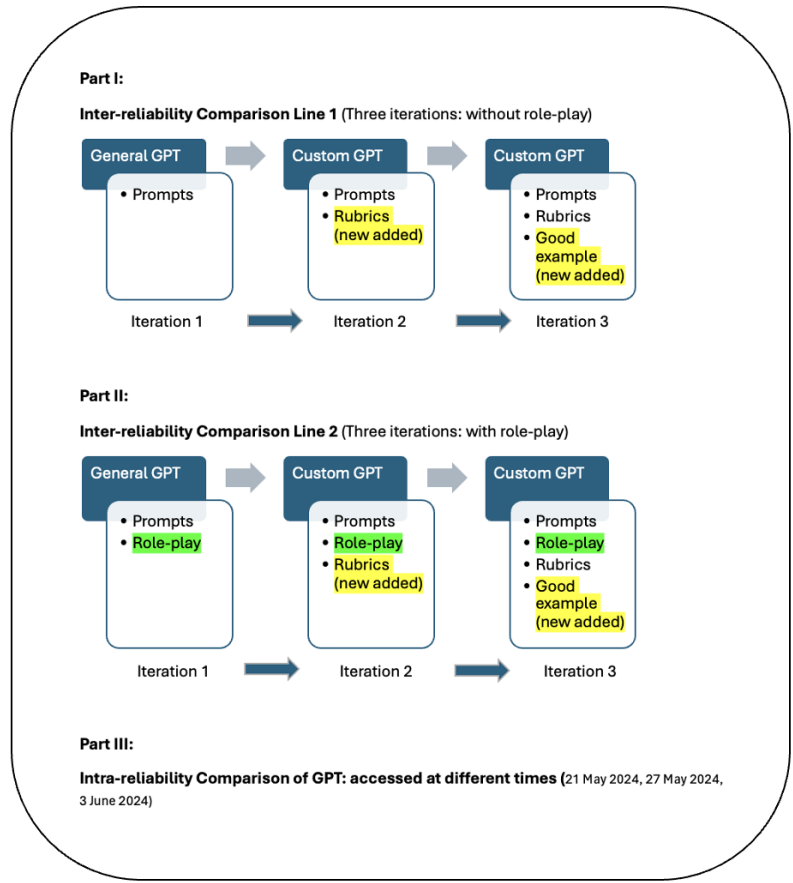

Figure 1. DBR approach in this study

As Figure 1 shows, there are three parts to this study. Part I is the 1st comparison line of the GPT's performance without role-playing. The 1st iteration uses General GPT to grade students' design assignments; the 2nd iteration uses ChatGPT-4 to create a Custom GPT, incorporating rubrics set by instructors; the 3rd iteration provides a good example in the Custom GPT, which is an assignment both instructors agree is good. It is well known in the literature, that giving examples of what is deemed as good output is a crucial element of prompting, that the GPT can then use as a benchmark against which they assess subsequent input provided. (Reynolds, & McDonell, 2021).

Part II is the second comparison line of GPT's performance, in which we have included role-play instructions, meaning that each iteration is evaluated from the perspectives of an instructor in a specific domain (in our case, architecture instructors and an engineering instructors). Specifically: the 1st iteration used General GPT for role-play grading; the 2nd iteration used Custom GPT for role-play grading (after adding Rubrics); the 3rd iteration used a Custom GPT with added rubrics and a good example for role-play.

Part III is the third comparison line assessing the intra-reliability of GPT itself. Here, we checked if the custom GPT evaluates a specific assignment consistently over time.

This study aims to answer two overarching research questions:

1) Can GPT accurately grade students' assignments in a design course, and 2) under which conditions can a Custom GPT act as a reliable grader to serve as a complementary rater?

> **Assumption 1:** In the iterations without role-play (Part I), inter-reliability increases progressively;
>
> **Assumption 2:** In the iterations with role-play (Part II), inter-reliability increases progressively;
>
> **Assumption 3:** When using GPT to score at different times (Part III), intra-reliability is greater than 0.5.

## 4. Findings

### 4.1 (Part I) Inter-reliability between GPT and human raters without role-playing

<u>1st iteration: General GPT</u>

*Process:*

We gave the following prompt to General GPT to score the students' assignments[3]. Because we need to compare the reliability with instructors' scoring, General GPT was given the same standards and scoring range as the instructors. However, detailed rubrics were not provided to General GPT in this iteration.

> *(Prompt given): Can you grade the following projects with the following three dimensions: 1) Design goal: whether it is appropriate, clear, and concise framing of the design goal from project, process, and skills perspectives (Full mark 15) 2) Site drawing: whether it shows at an appropriate level of abstraction with enough detail the key elements (Full mark 35) 3) Macro-AEIOU[4]: whether it provides an impactful sequence of visuals and comprehensive layers (Full mark 50)*

---

[3] The student's assignment is about designing for "Light." The student needs to identify where there are lighting issues in a designated area and then figure out how to design lights to solve the problem. For example, if a hallway in a shopping mall is particularly dark and few people pass through, the student identifies this issue and then designs an interactive light to increase people's interaction and engagement.
[4] AEIOU stands for: A – Activities; E – Environments; I – Interactions; O – Objects; U - Users

*Results:*

As shown in Table 1, the inter-reliability between the Architect Instructor and Engineer Instructor is not high, as they are scoring from two different perspectives. In this iteration, the inter-reliability between GPT and the two instructors is also not high. The inter-reliability between GPT and Engineer instructor slightly higher with the average scores of the two instructors at 0.4730, but still below 0.5.

Table 1. ICC Results of Iteration 1 (Part I without role-playing)

| Comparison | | ICC | Reliability | Explanation |
|---|---|---|---|---|
| **Architect Instructor** | Engineer instructor | 0.167 | Poor reliability | Two instructors differ a lot |
| **GPT** | Average of instructors | 0.2617 | Poor reliability | |
| **GPT** | Architect instructor | 0.1754 | Poor reliability | |
| **GPT** | Engineer instructor | 0.4730 | Poor reliability | |

2nd iteration: Custom GPT (with rubrics added)

*Process:*

Researchers used the rubrics provided by the instructors as prompts to build a Custom-GPT model. The building process included the following steps:

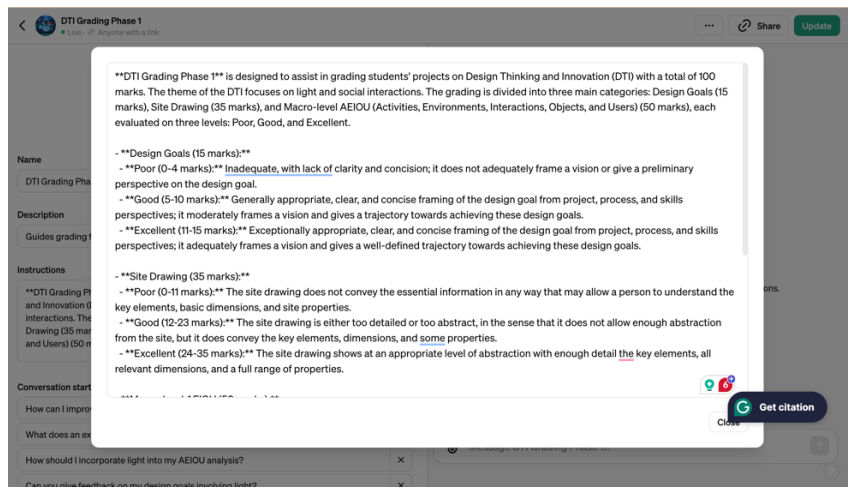

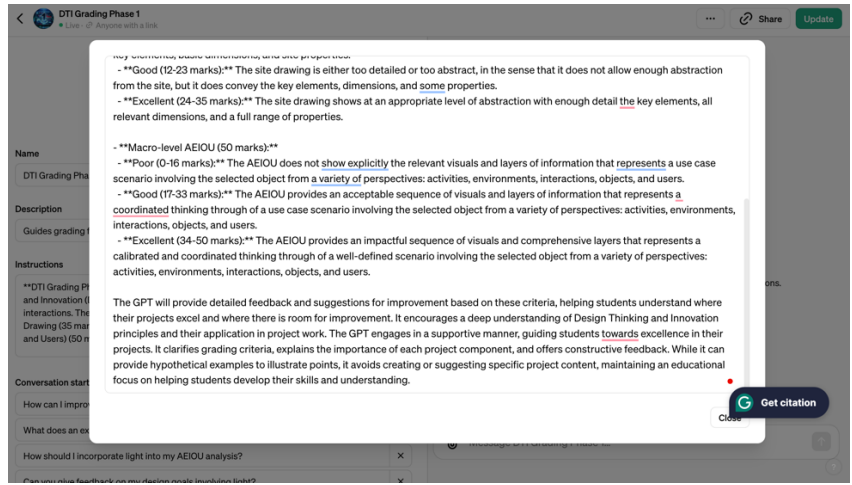

Screenshot 1. Prompts in building GPT for grading and feedback

After building the custom-GPT, the researchers took screenshots of each student's design assignment and fed these screenshots into the Custom-GPT one by one. The GPT was programmed to provide scores as well as feedback based on the rubrics for each student, including suggestions for their improvement.

*Results:*

In this iteration, the inter-reliability between GPT and the two instructors has improved. The ICC with the Architect Instructors increased by 0.2378, and with the Engineer Instructors by 0.1073, reaching 0.5803. At the same time, the inter-reliability with the average scores of the two instructors also improved by 0.3055, exceeding 0.5 (0.5672).

Table 2. ICC Results of Iteration 2 (Part I without role-playing)

| Comparison | | ICC | Reliability | Explanation |
| --- | --- | --- | --- | --- |
| **Architect instructor** | Engineer instructor | 0.167 | Poor reliability | Two instructors differ a lot |
| **GPT** | Average of instructors | 0.5672 | Moderate reliability | +0.3055 |
| **GPT** | Architect instructor | 0.4132 | Poor reliability | +0.2378 |
| **GPT** | Engineer instructor | 0.5803 | Moderate reliability | +0.1073 |

3rd iteration: Custom GPT (with a good example added)

*Process:*

The authors revised the instructions to the Custom-GPT as shown in the following screenshots, by giving it an example of what would be considered a good assignment.

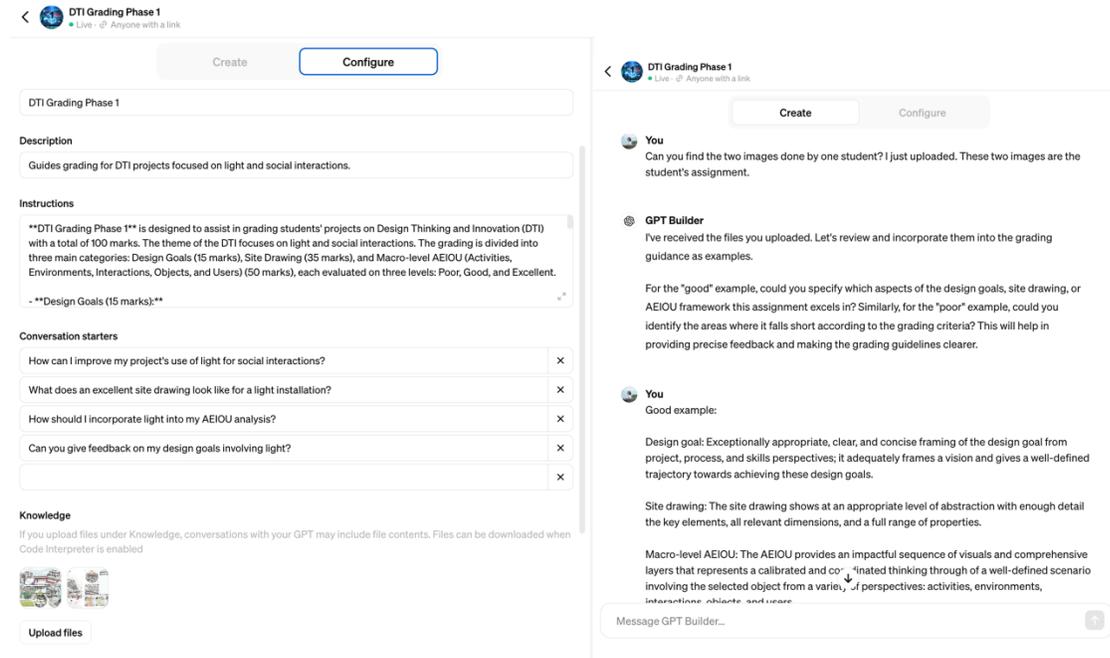

Screenshot 2. The bult-GPT was given one good example

*(Prompt given): Please rate the following assignments based on the rubrics given to you; and refer to the good example given to you.*

*Results:*

In this iteration, the inter-reliability between GPT and the two instructors continued to improve. The ICC with the Architect Instructors increased by 0.1018, and with the Engineer Instructors, it exceeded 0.7 (0.7652). At the same time, the inter-reliability with the average scores of the two instructors reached 0.7285. This already meets the requirements for raters' inter-reliability in high-stake exams.

Table 3. ICC Results of Iteration 3 (Part I without role-playing)

|  | Comparison | ICC | Reliability | Explanation |
|---|---|---|---|---|
| **Architect instructor** | Engineer instructor | 0.167 | Poor reliability | Two instructors differ a lot |
| **GPT** | Average of instructors | 0.7285 | High reliability | High +0.1613 |
| **GPT** | Architect instructor | 0.5150 | Moderate reliability | +0.1018 |
| **GPT** | Engineer instructor | 0.7652 | High reliability | High +0.1849 |

*\* All ICC data analysis in Appendix 1*

To sum up, the 4.1 section discussed the above answers the Assumptions 1 that in the iterations without role-play (Part I), inter-reliability increases progressively.

## 4.2 (Part II) Inter-reliability between GPT and human raters with role-play

### 1st Iteration: General GPT with Role-play

*Process:*

Based on the 1st iteration with General GPT, this time we let General GPT grade the students' assignments from the perspectives of both the architect instructor and the engineer instructor.

*Results:*

From the table below, it can be seen that the inter-reliability between Architect GPT and the Architect instructor is very low (0.1874), while the inter-reliability between Engineer GPT and the Engineer instructor is 0.5423 (Moderate reliability). The inter-reliability between the two GPTs (Architect GPT and Engineer GPT) is 0.5407 (Moderate reliability).

Table 4. ICC Results of Iteration 1 (Part II with role-playing)

| Comparison | | ICC | Reliability | Explanation |
|---|---|---|---|---|
| Architect GPT | Architect Instructor | 0.1874 | Poor reliability | |
| Engineer GPT | Engineer Instructor | 0.5423 | Moderate reliability | |
| Architecture GPT | Engineering GPT | 0.5407 | Moderate reliability | Two GPTs have higher reliability |

### 2nd iteration: Custom GPT (with rubrics added, with role-playing)

*Process:*

This iteration involved adding the rubrics used by the two instructors, and then conducting role-play to score from the perspectives of both the architect instructor and the engineer instructor.

*Results:*

The inter-reliability between Architect GPT and the Architect instructor increased by 0.2068, while the inter-reliability between Engineer GPT and the Engineer instructor decreased by 0.3024. The inter-reliability between the two GPTs (Architect GPT and Engineer GPT) decreased by 0.0132, but still exceeded 0.5. Therefore, in the next round, we will see if providing a good example to Custom-GPT will improve the reliability in each category.

Table 5. ICC Results of Iteration 2 (Part II with role-playing)

| Comparison | | ICC | Reliability | Explanation |
|---|---|---|---|---|
| Architect GPT | Architecture Instructor | 0.3942 | Poor reliability | +0.2068 |
| Engineer GPT | Engineer Instructor | 0.2399 | Poor reliability | -0.3024 |
| Architecture GPT | Engineering GPT | 0.5275 | Moderate reliability | -0.0132 |

Upon re-examining the details of the study, the fact that the Inter-Class Correlation (ICC) results in Table 4 (which uses general GPT) are higher than those in Table 5 (which uses custom GPT) indeed seems counterintuitive. The possible reasons include: The role-playing aspect adds another layer of complexity, as it requires the GPT model to simulate the perspectives of different types of instructors (architect and engineer). The custom GPT, being more finely tuned to specific rubric criteria, might struggle to adapt its grading process when switching between these varied instructional perspectives. Secondly, a "good example" may be needed for the GPT to reference. Without this example, the Custom GPT might still have had difficulty fully aligning its grading with the nuanced expectations of the human instructors.

3rd iteration: Custom GPT (with a good example added, with role-play)

*Process:*

This iteration included a good example, and then scoring was conducted from the perspectives of both the architect instructor and the engineer the instructor.

*Results:*

Table 6. ICC Results of Iteration 3 (Part II with role-playing)

| Comparison | | ICC | Reliability | Explanation |
|---|---|---|---|---|
| Architect GPT | Architecture Instructor | 0.7199 | High reliability | +0.3257 |
| Engineer GPT | Engineer Instructor | 0.5554 | Moderate reliability | +0.3155 |
| Architecture GPT | Engineering GPT | 0.5536 | Moderate reliability | +0.0261 |

*ICC data analysis in this iteration in Appendix 2

To sum up, this section discussed the above answers the Assumption 2 that in the iterations with role-play (Part II), inter-reliability increases progressively. Furthermore, in each iteration in role-playing, the inter-reliability between the Architecture GPT and the Engineering GPT is above 0.5 while the inter-reliability between the human Architecture instructor and the human Engineering instructors is only 0.167.

### 4.3 (Part III) Intra-reliability of GPT accessed in three different times

To verify the intra-reliability of GPT's scoring over different times, researchers randomly conducted three comparisons on different days, using General GPT to grade the same student assignments.

| Project No. | 1st time<br>Accessed on 21 May 2024 | 2nd time<br>Accessed on 27 May 2024 | 3rd time<br>Accessed on 3 June |
|---|---|---|---|
| 1 | 89 | 92 | 89 |
| 2 | 94 | 93 | 93 |
| 3 | 82 | 78 | 79 |
| 4 | 95 | 95 | 93 |
| 5 | 92 | 92 | 95 |
| 6 | 95 | 94 | 93 |
| 7 | 95 | 95 | 93 |
| 8 | 93 | 91 | 91 |
| 9 | 92 | 93 | 94 |
| 10 | 91 | 93 | 94 |

The ICC among these 3 iteration is 0.6485. The ICC between 1$^{st}$ and 2$^{nd}$: 0.7336. The ICC between 2$^{nd}$ and 3$^{rd}$ is 0.7750. The ICC between 1$^{st}$ and 3$^{rd}$: 0.6983. Thus, the intra-reliability of GPT is high.

To sum up, this section discussed the above answers the Assumption 3 that when using GPT to score at different times (Part III), intra-reliability is greater than 0.5. The 4.1 through 4.3 sections answered RQ1 by verifying the three assumptions

4.4 Custom GPT's role in providing feedback

Besides, while GPT was grading students' assignment, it provided with personalized feedback, no matter without or with role-playing. As Screenshot 4 shows, the Custom-GPT knows the perspectives from both engineer instructor and Architect instructor. It provides feedback from different perspectives so that students can receive the different opinions and balance different feedback in a more creative way (Screenshot 3).

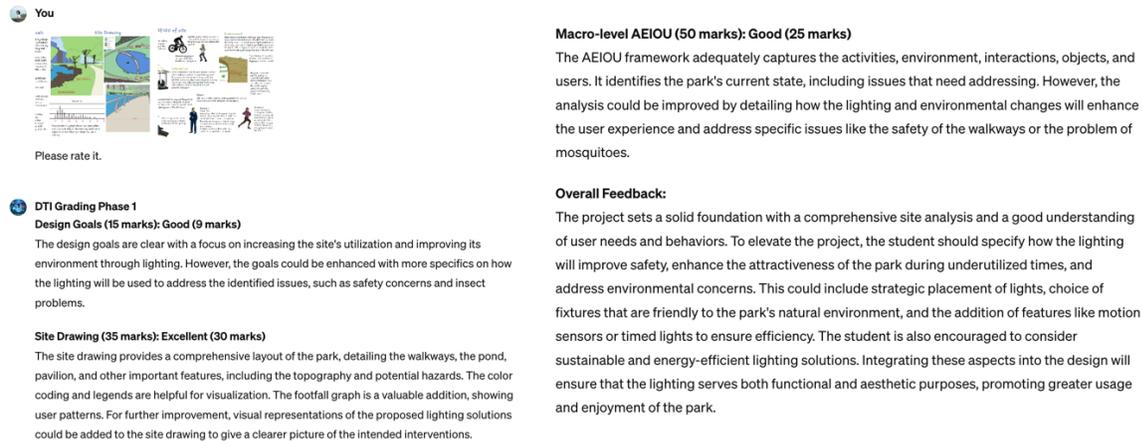

Screenshot 3. Custom-GPT in grading and providing feedback

Thus, GPT-4 is an effective tool for providing students with timely feedback, as instructors lack the capacity to: 1) provide individual feedback to a large number of students; 2) offer advice to students at all times (7 days a week, 24 hours a day). For design students, GPT can make a significant contribution to providing timely feedback to students, and it is worth considering its use on a large scale to continuously encourage students to promptly improve their design assignments and provide students with synthesis of instructors from different background – in this research, engineer instructor and architect instructors' perspective (shown in Screenshot 5). At the meantime, GPT can could serve as a reference or a co-worker for instructors. For instance, if instructors find a significant discrepancy in the scores of a particular student, they can focus on why this discrepancy occurs, and even question whether the rubrics were scientifically set.

**Synthesis of Both Perspectives:**

Combining the **engineer's practical focus on functionality and safety** with the **architect's attention to spatial aesthetics and user experience** would result in a stronger project. Key improvements include:

1. **Integrating Technical and Aesthetic Aspects**: Address **lighting efficiency** and **maintenance** from an engineering viewpoint, while also considering **how light enhances spatial quality** from an architectural perspective.
2. **Detailed Drawings**: Include **site plans** and **elevations** with both **technical details** (e.g., wiring, dimensions) and **aesthetic considerations** (e.g., light placement and interaction with materials).
3. **Comprehensive AEIOU Analysis**: Combine **user safety and functionality** (engineer's perspective) with **user experience and comfort** (architect's perspective).

With these improvements, the project could better balance **functionality, safety, and aesthetics**, leading to a more comprehensive design.

Screenshot 4. Custom-GPT provided feedback by making a synthesis of both engineer and architect instructors' perspective

Furthermore, Custom-GPT can be given prompts to compare students' work with that of a renowned designer, providing directions and suggestions for emulation. This approach could involve GPT analyzing notable works in the field of design and drawing parallels between these works and the students' assignments. GPT could then offer constructive feedback on how students might incorporate certain elements or techniques from established designers into their own work to improve creativity, functionality, and aesthetic appeal. This comparative analysis could serve as an inspirational and educational tool, helping students understand the benchmarks in their field and aspire to reach or exceed them. For instance, Screenshot 8 shows the comparison between Teamlab and students' project and GPT provided further suggestions on improvement to students.

> *(Prompt given): Can compare this design project with the renowned Teamlab? any suggestions?*

Screenshot 5. Custom-GPT compares student's design project with Teamlab

Thus, during GPT's grading process, it has many other roles: Firstly, it can provide 24-hour/day and 7-days/week personalized feedback to students. Secondly, through personalized feedback and being given prompts, it can stimulate students' ideas and inspirations. Thirdly, it can serve as a co-worker with instructors to provide references during instructors' grading process.

## Discussion

This study demonstrates through two lines of iteration how GPT can effectively grade students' design assignments. We found that with the appropriate instruction the evaluations of custom GPTs can have high inter-reliability and intra-reliability. Using educational design-based research, this study provides a detailed demonstration of each iteration's process and offers the following generalized principle guidelines:

1) Provide GPT with accurate and consistent prompts;

2) Establish Custom-GPT with rubrics consistent with those used by instructors;

3) Give Custom-GPT a good example to clarify what constitutes a good project/assignment.

Thus, this study verifies using ChatGPT to mimic human raters in scoring and normalizing grading processes can achieve high inter-reliability with human raters. The implications of our findings indicate that GPT, under the right set of conditions, can serve as a co-worker to instructors in educational settings (Mollick, 2023). In terms of grading, both intra-rater and inter-rater reliability are two key rules for the validity and fairness of educational assessments. The studies reviewed herein underscore the importance of structured rater training, the use of clear rubrics, and the necessity for regular calibration and feedback mechanisms. As educational methods continue to evolve, so too must the approaches to ensuring that assessments are reliable and equitable. GPT models offer promising tools to support this process but must be integrated thoughtfully to avoid introducing new biases and complexities. Continuing to innovate in the tools and methods used to train and calibrate raters will be essential in maintaining the integrity and fairness of educational evaluations.

On the one hand, GPT models can support the process by providing consistent, unbiased scoring guidelines and assisting in the training and calibration of human raters. For example, automated feedback and analysis provided by GPT models can help raters understand and correct their biases, leading to improved reliability over time. On the other hand, the introduction of GPT models brings new complexities into the equation. There are concerns about the transparency of AI-driven decisions and the potential for the technology to perpetuate existing biases if not properly monitored. Additionally, reliance on AI tools may lead to reduced critical engagement by human raters, potentially affecting the overall quality of the assessments.

Besides, as our findings show, the feedback provided by Custom-GPT can promote personalized learning for students. It aligns with insights from recent studies. One study, for instance, found that ChatGPT can promote personalized and interactive learning, particularly in generating prompts for formative assessment and ongoing feedback, although concerns exist regarding the potential for generating incorrect information and inherent biases (Baidoo-Anu & Owusu Ansah, 2023). ChatGPT has demonstrated efficacy in providing detailed feedback on language learning, with its evaluations closely aligning with those of experienced instructors (Dai et al., 2023). From this study, GPT can serve as a co-worker for educators to increase productivity.

## Conclusion

This research has provided instructors and educators with insights on how to better utilize technology to assist in grading. Firstly, it emphasizes the need to embrace technology as a means to enhance work efficiency through interaction between humans and technology. Secondly, the study illustrates the iterative process of building a Custom-GPT in evaluating design students' assignments by showing hands-on empirical process. It offers practically meaningful prompts and process in using GPT-4 as co-workers. Thirdly, the study highlights the need for more empirical studies in demonstrating the impact of using GPT in design students' learning processes.

Zhu, H., & Wang, Y. (2023). *Using ChatGPT for academic writing support: Error detection and improvement*. *Journal of Educational Computing Research, 61*(1), 45-62. https://doi.org/10.2190/EC.61.1.a

## Appendix 1

ICC results of the 5th iteration (without role-play):

**Anova: Two-Factor Without Replication**

| SUMMARY | Count | Sum | Average | Variance |
|---|---|---|---|---|
| Row 1 | 2 | 147 | 73.5 | 40.5 |
| Row 2 | 2 | 171 | 85.5 | 24.5 |
| Row 3 | 2 | 110 | 55 | 2 |
| Row 4 | 2 | 181 | 90.5 | 24.5 |
| Row 5 | 2 | 176 | 88 | 2 |
| Row 6 | 2 | 173 | 86.5 | 12.5 |
| Row 7 | 2 | 170 | 85 | 72 |
| Row 8 | 2 | 171 | 85.5 | 112.5 |
| Row 9 | 2 | 176 | 88 | 0 |
| Row 10 | 2 | 180 | 90 | 18 |
| | | | | |
| Column 1 | 10 | 831 | 83.1 | 125.211111 |
| Column 2 | 10 | 824 | 82.4 | 143.822222 |

**ANOVA**

| Source of Variation | SS | df | MS | F | P-value | F crit |
|---|---|---|---|---|---|---|
| Rows | 2115.25 | 9 | 235.027778 | 6.91145238 | 0.00410966 | 3.1788931 |
| Columns | 2.45 | 1 | 2.45 | 0.07204705 | 0.79443034 | 5.11735503 |
| Error | 306.05 | 9 | 34.0055556 | | | |
| | | | | | | |
| Total | 2423.75 | 19 | | | | |

0.76515119

| Anova: Two-Factor Without Replication | | | | | | |
|---|---|---|---|---|---|---|
| SUMMARY | Count | Sum | Average | Variance | | |
| Row 1 | 2 | 134 | 67 | 8 | | |
| Row 2 | 2 | 159.5 | 79.75 | 10.125 | | |
| Row 3 | 2 | 99 | 49.5 | 84.5 | | |
| Row 4 | 2 | 172 | 86 | 128 | | |
| Row 5 | 2 | 168 | 84 | 50 | | |
| Row 6 | 2 | 164 | 82 | 8 | | |
| Row 7 | 2 | 167.5 | 83.75 | 105.125 | | |
| Row 8 | 2 | 160 | 80 | 8 | | |
| Row 9 | 2 | 173.5 | 86.75 | 3.125 | | |
| Row 10 | 2 | 174 | 87 | 72 | | |
| | | | | | | |
| Column 1 | 10 | 747.5 | 74.75 | 153.013889 | | |
| Column 2 | 10 | 824 | 82.4 | 143.822222 | | |
| | | | | | | |
| ANOVA | | | | | | |
| Source of Variation | SS | df | MS | F | P-value | F crit |
| Rows | 2487.2625 | 9 | 276.3625 | 13.4984736 | 0.00032137 | 3.1788931 |
| Columns | 292.6125 | 1 | 292.6125 | 14.2921783 | 0.00434567 | 5.11735503 |
| Error | 184.2625 | 9 | 20.4736111 | | | |
| | | | | | | |
| Total | 2964.1375 | 19 | | | | |

0.72848049

# Appendix 2

ICC results of the 6th iteration (without role-play):

| Anova: Two-Factor Without Replication | | | | | | |
|---|---|---|---|---|---|---|
| SUMMARY | Count | Sum | Average | Variance | | |
| Row 1 | 2 | 118 | 59 | 98 | | |
| Row 2 | 2 | 131 | 65.5 | 0.5 | | |
| Row 3 | 2 | 85 | 42.5 | 220.5 | | |
| Row 4 | 2 | 146 | 73 | 32 | | |
| Row 5 | 2 | 142 | 71 | 0 | | |
| Row 6 | 2 | 141 | 70.5 | 0.5 | | |
| Row 7 | 2 | 148 | 74 | 0 | | |
| Row 8 | 2 | 148 | 74 | 18 | | |
| Row 9 | 2 | 161 | 80.5 | 12.5 | | |
| Row 10 | 2 | 148 | 74 | 2 | | |
| | | | | | | |
| Column 1 | 10 | 664 | 66.4 | 207.6 | | |
| Column 2 | 10 | 704 | 70.4 | 57.3777778 | | |
| | | | | | | |
| ANOVA | | | | | | |
| Source of Variation | SS | df | MS | F | P-value | F crit |
| Rows | 2080.8 | 9 | 231.2 | 6.84473684 | 0.00425482 | 3.1788931 |
| Columns | 80 | 1 | 80 | 2.36842105 | 0.15819542 | 5.11735503 |
| Error | 304 | 9 | 33.7777778 | | | |
| | | | | | | |
| Total | 2464.8 | 19 | | | | |

0.71993517

# Appendix 3

Grading results of two human instructors.

|  | Architect Instructor | | | | Engineer Instructor | | | | |
| --- | --- | --- | --- | --- | --- | --- | --- | --- | --- |
| Project No. | Design Goal | Site Drawing | Macro-AEIOU | Total | Design Goal | Site Drawing | Macro-AEIOU | Total | Average |
| 1 | 9 | 15 | 28 | 52 | 14 | 24 | 40 | 78 | 65 |
| 2 | 11 | 25 | 30 | 66 | 14 | 32 | 43 | 89 | 77.5 |
| 3 | 4 | 12 | 16 | 32 | 10 | 13 | 31 | 54 | 43 |
| 4 | 11 | 23 | 35 | 69 | 14 | 32 | 41 | 87 | 78 |
| 5 | 11 | 25 | 35 | 71 | 14 | 32 | 41 | 87 | 79 |
| 6 | 11 | 25 | 35 | 71 | 14 | 30 | 45 | 89 | 80 |
| 7 | 10 | 31 | 33 | 74 | 14 | 31 | 34 | 79 | 76.5 |
| 8 | 10 | 25 | 36 | 71 | 14 | 32 | 47 | 93 | 82 |
| 9 | 11 | 32 | 40 | 83 | 14 | 33 | 41 | 88 | 85.5 |
| 10 | 11 | 30 | 34 | 75 | 14 | 34 | 39 | 87 | 81 |

## Appendix 4

Grading results of each iteration.

|  | 1st Iteration: General GPT | | | |
| --- | --- | --- | --- | --- |
| Project No. | Design Goal | Site Drawing | Macro-AEIOU | Total |
| 1 | 14 | 30 | 45 | 89 |
| 2 | 14 | 33 | 47 | 94 |
| 3 | 14 | 28 | 40 | 82 |
| 4 | 14 | 33 | 48 | 95 |
| 5 | 14 | 32 | 46 | 92 |
| 6 | 15 | 32 | 48 | 95 |
| 7 | 15 | 33 | 47 | 95 |
| 8 | 15 | 32 | 46 | 93 |
| 9 | 14 | 32 | 46 | 92 |
| 10 | 14 | 32 | 45 | 91 |

|  | 2nd Iteration: General GPT (Engineers) | | | | 2nd Iteration: General GPT (Architects) | | | |
| --- | --- | --- | --- | --- | --- | --- | --- | --- |
| Project No. | Design Goal | Site Drawing | Macro-AEIOU | Total | Design Goal | Site Drawing | Macro-AEIOU | Total |
| 1 | 12 | 28 | 40 | 80 | 14 | 32 | 45 | 91 |
| 2 | 15 | 35 | 50 | 100 | 15 | 35 | 50 | 100 |
| 3 | 12 | 30 | 42 | 84 | 12 | 28 | 40 | 80 |
| 4 | 14 | 32 | 45 | 91 | 13 | 30 | 47 | 90 |
| 5 | 13 | 28 | 40 | 81 | 14 | 32 | 45 | 91 |
| 6 | 15 | 30 | 45 | 90 | 15 | 32 | 48 | 95 |
| 7 | 13 | 30 | 45 | 88 | 14 | 32 | 48 | 94 |
| 8 | 12 | 28 | 42 | 82 | 13 | 30 | 45 | 88 |
| 9 | 13 | 30 | 45 | 88 | 14 | 32 | 48 | 94 |
| 10 | 15 | 30 | 45 | 90 | 15 | 32 | 48 | 95 |

|  | 3rd Iteration: General GPT | | | |
|---|---|---|---|---|
| Project No. | Design Goal | Site Drawing | Macro-AEIOU | Total |
| 1 | 13 | 30 | 45 | 88 |
| 2 | 9 | 28 | 30 | 67 |
| 3 | 9 | 22 | 25 | 56 |
| 4 | 13 | 30 | 45 | 88 |
| 5 | 14 | 32 | 45 | 91 |
| 6 | 14 | 32 | 45 | 91 |
| 7 | 14 | 32 | 45 | 91 |
| 8 | 11 | 30 | 33 | 74 |
| 9 | 13 | 32 | 45 | 90 |
| 10 | 14 | 32 | 46 | 92 |

|  | 4th Iteration: General GPT (Engineers) | | | | 4th Iteration: General GPT (Architects) | | | |
|---|---|---|---|---|---|---|---|---|
| Project No. | Design Goal | Site Drawing | Macro-AEIOU | Total | Design Goal | Site Drawing | Macro-AEIOU | Total |
| 1 | 12 | 23 | 35 | 70 | 13 | 25 | 44 | 82 |
| 2 | 11 | 25 | 24 | 60 | 12 | 28 | 32 | 72 |
| 3 | 13 | 25 | 27 | 65 | 11 | 23 | 28 | 62 |
| 4 | 14 | 28 | 42 | 84 | 14 | 30 | 45 | 89 |
| 5 | 13 | 29 | 38 | 80 | 14 | 32 | 46 | 92 |
| 6 | 10 | 28 | 35 | 73 | 9 | 30 | 42 | 81 |
| 7 | 11 | 27 | 43 | 81 | 13 | 32 | 45 | 90 |
| 8 | 11 | 27 | 34 | 72 | 13 | 32 | 45 | 90 |
| 9 | 12 | 29 | 38 | 79 | 14 | 33 | 46 | 93 |
| 10 | 13 | 30 | 40 | 83 | 14 | 34 | 48 | 96 |

|  | 5th Iteration: General GPT |
|---|---|

| Project No. | Design Goal | Site Drawing | Macro-AEIOU | Total |
|---|---|---|---|---|
| 1 | 11 | 24 | 34 | 69 |
| 2 | 13 | 29 | 40 | 82 |
| 3 | 9 | 19 | 28 | 56 |
| 4 | 14 | 33 | 47 | 94 |
| 5 | 14 | 30 | 45 | 89 |
| 6 | 13 | 29 | 42 | 84 |
| 7 | 13 | 32 | 46 | 91 |
| 8 | 12 | 27 | 39 | 78 |
| 9 | 13 | 30 | 45 | 88 |
| 10 | 14 | 32 | 47 | 93 |

| | 6th Iteration: General GPT (Engineers) | | | | 6th Iteration: General GPT (Architects) | | | |
|---|---|---|---|---|---|---|---|---|
| Project No. | Design Goal | Site Drawing | Macro-AEIOU | Total | Design Goal | Site Drawing | Macro-AEIOU | Total |
| 1 | 11 | 28 | 39 | 78 | 10 | 24 | 32 | 66 |
| 2 | 13 | 26 | 33 | 72 | 11 | 24 | 30 | 65 |
| 3 | 12 | 20 | 30 | 62 | 10 | 18 | 25 | 53 |
| 4 | 12 | 25 | 34 | 71 | 13 | 27 | 37 | 77 |
| 5 | 13 | 28 | 37 | 78 | 12 | 25 | 34 | 71 |
| 6 | 13 | 27 | 38 | 78 | 12 | 25 | 33 | 70 |
| 7 | 13 | 30 | 40 | 83 | 12 | 27 | 35 | 74 |
| 8 | 14 | 30 | 42 | 86 | 13 | 28 | 36 | 77 |
| 9 | 14 | 30 | 41 | 85 | 13 | 28 | 37 | 78 |
| 10 | 13 | 29 | 40 | 82 | 12 | 26 | 35 | 73 |

**Appendix 5**

Screenshots of grading a student's project from an engineer's perspective and an architecture's perspective.

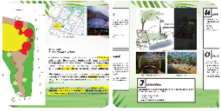

**Grading from an Engineer's Perspective:**

1. Design Goals (15 marks)

- **Score: 12/15 (Excellent)**
- **Feedback:** The design goal is clear and oriented towards practical enhancements, focusing on improving lighting to extend the usability of the garden into the night. It appropriately addresses the current limitations and aims to enhance user interaction and safety through light installations. Incorporating specific engineering solutions such as the types of lights and energy-efficient measures could enrich the proposal.

2. Site Drawing (35 marks)

- **Score: 25/35 (Good)**
- **Feedback:** The site drawings clearly delineate the layout and intended additions like benches and lighting. The representation of elevation changes and spatial planning is practical and informative. For improved engineering accuracy, inclusion of details regarding the infrastructure needed for lighting installations, like electrical routing and structural supports, would be beneficial.

3. Macro-level AEIOU (50 marks)

- **Score: 38/50 (Excellent)**
- **Feedback:** The AEIOU analysis successfully covers various aspects of the project, from user interactions to environmental impact. The detailed descriptions of activities and objects in the garden demonstrate a good understanding of how the space is utilized. Including more technical insights into how proposed engineering changes impact these elements could further strengthen the evaluation.

**Total Score: 75/100**

**Grading from an Architect's Perspective:**

1. Design Goals (15 marks)

- **Score: 13/15 (Excellent)**
- **Feedback:** The design goals effectively reflect an architectural vision that prioritizes aesthetic enhancement and user experience. The focus on light installations to improve night-time ambiance and encourage interactions is commendable. Expanding on how these installations will integrate with the garden's natural elements could provide a more comprehensive design narrative.

2. Site Drawing (35 marks)

- **Score: 28/35 (Excellent)**
- **Feedback:** The drawings effectively communicate the spatial arrangement and proposed enhancements. They visually support the goal of transforming the area into a more inviting and functional space. Further elaboration on material choices and architectural details such as the design of light fixtures and benches would enhance the presentation.

3. Macro-level AEIOU (50 marks)

- **Score: 42/50 (Excellent)**
- **Feedback:** This section is well executed, showing a deep understanding of the site's dynamics, including user behavior and environmental conditions. Architecturally, incorporating design principles that focus on sustainability and user comfort, such as solar-powered lighting or adaptive landscape features, could enrich the project.

**Total Score: 83/100**

---

**Overall Feedback:**

The project shows a strong conceptual understanding from both engineering and architectural perspectives, focusing on enhancing user experience and functionality through light installations. Both evaluations suggest that further detailing in technical specifications and design integration could elevate the project, ensuring that it not only meets functional needs but also enhances the aesthetic and environmental quality of the space. Incorporating sustainable practices and detailed architectural elements would significantly strengthen the final proposal.